\documentclass[runningheads]{llncs}
\usepackage[T1]{fontenc}
\usepackage{graphicx}
\usepackage{booktabs}
\usepackage{amsmath,amssymb,amsfonts}

\begin{document}
\title{Contrastive Learning through Auxiliary Branch for Video Object Detection}
%
%
\author{Lucas Rakotoarivony\orcidID{0009-0008-1383-6460}}
\authorrunning{L. Rakotoarivony}
%
\institute{Thales, cortAIx Labs \\ 
1 Av. Augustin Fresnel, 91120, Palaiseau, French \\
\email{lucas.rakotoarivony@thalesgroup.com}\\}
\maketitle              
\begin{abstract}
Video object detection is a challenging task because videos often suffer from image deterioration such as motion blur, occlusion, and deformable shapes, making it significantly more difficult than detecting objects in still images. Prior approaches have improved video object detection performance by employing feature aggregation and complex post-processing techniques, though at the cost of increased computational demands. To improve robustness to image degradation without additional computational load during inference, we introduce a straightforward yet effective Contrastive Learning through Auxiliary Branch (CLAB) method. First, we implement a constrastive auxiliary branch using a contrastive loss to enhance the feature representation capability of the video object detector’s backbone. Next, we propose a dynamic loss weighting strategy that emphasizes auxiliary feature learning early in training while gradually prioritizing the detection task as training converges. We validate our approach through comprehensive experiments and ablation studies, demonstrating consistent performance gains. Without bells and whistles, CLAB reaches a performance of 84.0\% mAP and 85.2\% mAP with ResNet-101 and ResNeXt-101, respectively, on the ImageNet VID dataset, thus achieving state-of-the-art performance for CNN-based models without requiring additional post-processing methods.

\end{abstract}
\section{Introduction}

Video object detection is a key area of computer vision that focuses on detecting objects in a continuous sequence of images. Unlike object tracking \cite{cui2022mixformer,zhang2021bytetrack}, which assumes known object locations and follows them across frames, video object detection requires detecting objects from scratch in each frame, making it inherently more challenging. Directly applying state-of-the-art detectors \cite{ren2015faster} to solve this task leads to suboptimal performance due to typical video challenges like motion blur, occlusion, and deformable shapes. To address these issues, previous works \cite{wu2019sequence,chen2020memory} have used images from the same video sequence, called "support frames", to extract relevant features and temporal context, which improves the detection in the target frame. However, to keep enhancing video object detectors, state-of-the-art methods rely on complex techniques to efficiently aggregate support frame knowledge. These include using external memory \cite{chen2020memory} or post-processing methods \cite{han2016seq,deng2019relation}, which require significant computational resources.

\begin{figure}[t]
\centerline{\includegraphics[width=1\columnwidth]{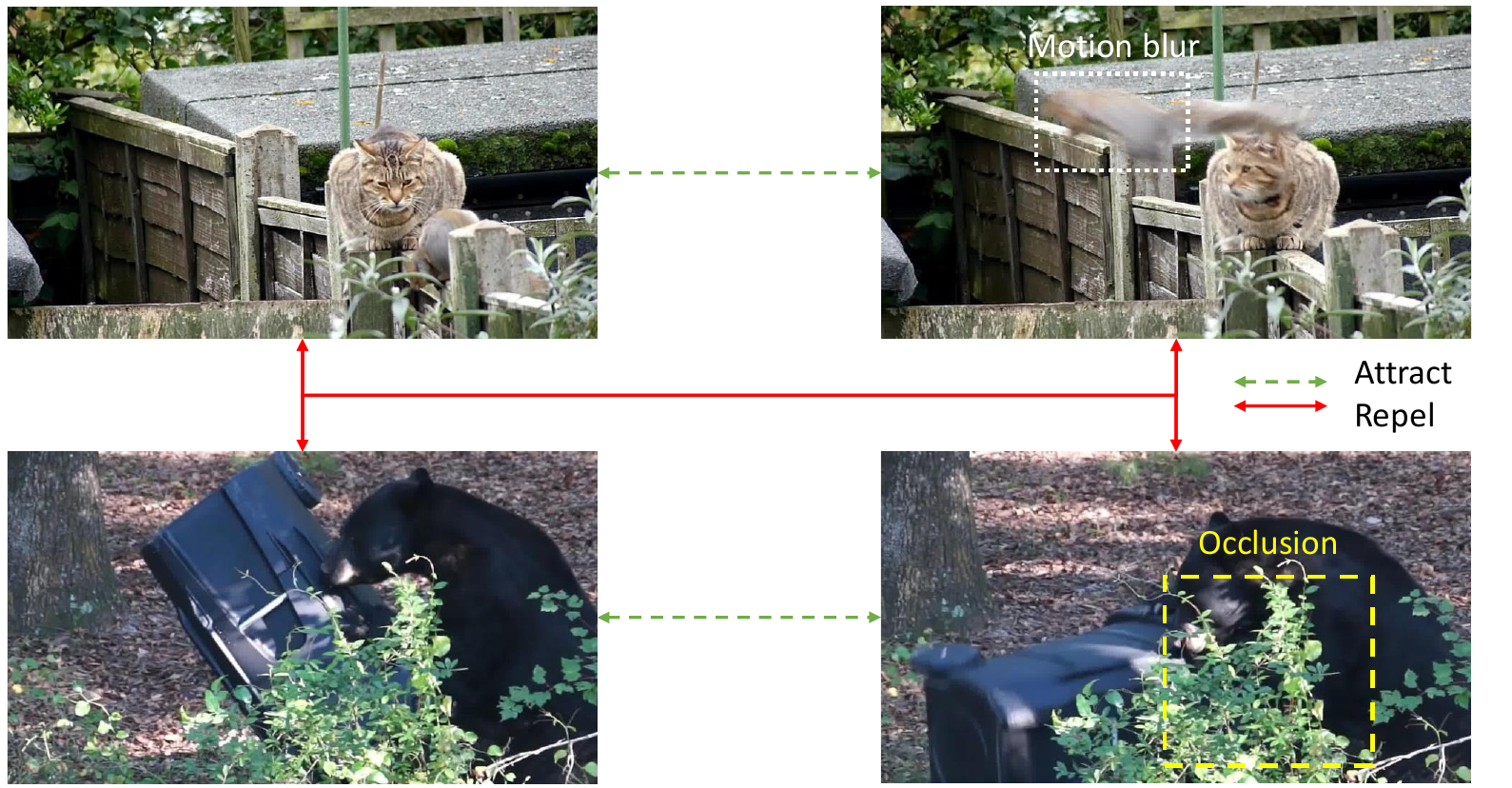}}
\caption{Illustration of the proposed contrastive auxiliary task. This task benefits the model in two key ways. First, by pulling together frames from the same video, the model becomes resilient to visual degradations like motion blur caused by rapid camera movements, as seen with the squirrel. Second, by pushing apart frames from different videos, the model learns generalizable features, such as global context, which help address challenges like the occlusion of the bear.}
\label{benefit}
\end{figure}

On the other hand, video representation learning methods \cite{qian2021spatiotemporal,dave2022tclr} focus on extracting rich, high-quality visual features from video clips. These methods aim to capture not only the spatial details of individual frames but also the temporal dynamics across frames, which is crucial for understanding motion, object interactions, and scene changes in videos. One of the most effective techniques in this task is contrastive learning \cite{oord2018representation,chen2020simple,he2020momentum}, which enhances the ability of a model to distinguish between different features within a video. By learning to pull together similar frames (positive pairs) and push apart dissimilar ones (negative pairs), contrastive learning helps the model develop more robust and discriminative feature representations. These representations are essential for downstream tasks like object detection, action recognition, and tracking, as they help in distinguishing objects even under challenging conditions mentioned previously.

To enhance the performance of video object detectors without adding computational complexity during inference, we propose a simple yet effective Contrastive Learning through Auxiliary Branch strategy dubbed as CLAB. This approach uses a constrastive auxiliary branch (CAB) based on the InfoNCE \cite{oord2018representation,chen2020simple} loss to improve the feature representations of a video clip. The benefits of this contrastive task are illustrated in Figure \ref{benefit}. In a batch of video sequences, we treat all images from the same video (i.e., the target frame and its support frames) as positive pairs, and images from different video sequences as negative pairs. Since this auxiliary branch is only used during training, it doesn’t increase computational costs during inference.

Additionally, inspired by the success of auxiliary strategies like \cite{lin2019adaptive,du2018adapting}, we introduce a dynamic loss weight (DLW) strategy. Instead of using a fixed weight to combine the auxiliary loss with the detection loss, DLW gradually decreases the loss weight to zero over the course of training. This approach allows the auxiliary branch to have a greater impact during the early stages, where learning general features is crucial, and ensures the model focuses on the primary task as training concludes.

We implemented our method on top of the Sequence Level Semantics Aggregation (SELSA) \cite{wu2019sequence} + Temporal ROI Align (TROI) \cite{gong2021temporal} method, leveraging SELSA’s long-range temporal information and TROI’s high-quality proposals. As these methods are based on a Faster R-CNN \cite{ren2015faster} model with a ResNet \cite{he2016deep} backbone, we evaluated our method with the same architecture and demonstrated its effectiveness in boosting performance on the popular ImageNet VID \cite{russakovsky2015imagenet} dataset without adding complexity at inference. ImageNet VID includes a wide range of target object sizes, such as small objects like squirrels, medium ones like cats, and large ones like bears, as illustrated in Figure \ref{benefit}, which highlights the value of improving feature representations under such variation.

Our paper’s main contributions can be summarized as follows:

\begin{itemize}
\item We propose a simple yet effective method, CLAB, which uses an auxiliary branch based on contrastive loss to enhance feature representation quality and, consequently, improve the performance of video object detectors.
\item We introduce DLW, an auxiliary loss weighting strategy that enables the model to effectively incorporate knowledge from the auxiliary branch without disrupting the primary task.
\item We evaluate CLAB on the widely used ImageNet VID video detection dataset, achieving 84.0\%  mAP with ResNet-101 and 85.2\%  mAP with ResNeXt-101, achieving state-of-the-art performance for CNN-based video object detectors without increasing computational complexity during inference.
\item We conduct extensive experiments to analyze our method’s performance under various hyperparameters.
\end{itemize}

\section{Related Work}

\subsection{Video Object Detection}

Video object detection is more challenging than still-image detection due to image degradation caused by motion blur, defocus, occlusion, and unusual object poses. Early video object detection methods combined motion and frame features, using optical flow \cite{zhu2017flow} to aggregate information and object tracking \cite{wang2018fully} to refine detection. \cite{Liu_2023_ICCV} improved accuracy by focusing on fewer frames, while \cite{bertasius2018object} employed deformable convolutions for pixel-level changes. Plug-and-play modules \cite{hashmi2023boxmask,gong2021temporal} adapted still-image components like ROI Align for video tasks. 

Most video object detection models rely on R-CNN-based architectures \cite{ren2015faster}, but their performance depends on the region proposal network (RPN), which can miss objects or generate false positives. However, Transformer-based architectures \cite{wang2022ptseformer,roh2023diffusionvid} are gaining popularity, leveraging long-range dependencies for better performance but at higher computational cost.

\subsection{Video Contrastive Self-supervised Learning}

Following the success of contrastive learning in self-supervised image representation \cite{chen2020simple,he2020momentum}, many methods extended it to video. Video contrastive self-supervised learning \cite{qian2021spatiotemporal} typically used instance-level discrimination, treating clips from the same video as positive pairs and those from different videos as negatives. \cite{zhuang2020unsupervised} combined 2D CNN static features with 3D CNN motion features, while \cite{qian2021spatiotemporal} refined temporal augmentation. \cite{dave2022tclr} introduced local and global temporal losses to enhance feature diversity.

\subsection{Auxiliary Learning}

Auxiliary learning enhances a primary task by leveraging auxiliary tasks to develop generalizable features, benefiting fields like reinforcement learning \cite{lin2019adaptive}, knowledge distillation \cite{xu2020knowledge}, transfer learning \cite{tzeng2015simultaneous}, and computer vision \cite{zhao2017pyramid}. In semantic segmentation, \cite{zhao2017pyramid} used an auxiliary branch to mitigate vanishing gradients, while \cite{araslanov2020single} explored joint classification and segmentation. Most methods optimize a weighted sum of primary and auxiliary losses \cite{zhai2019s4l}, though recent work \cite{lin2019adaptive} introduced adaptive weighting.

\section{Methods}

\textbf{Overview.} In this section, we present CLAB, a simple yet effective method that enhances video object detection performance without adding computational overhead during inference. Our approach introduces a constrastive auxiliary branch that applies the InfoNCE \cite{oord2018representation,he2020momentum,chen2020simple} loss on video sequence features, designating frames from the same video as positive pairs and those from different videos as negatives to improve robustness against frame degradation. 

\begin{figure*}[t]
\centering
\includegraphics[width=0.85\textwidth]{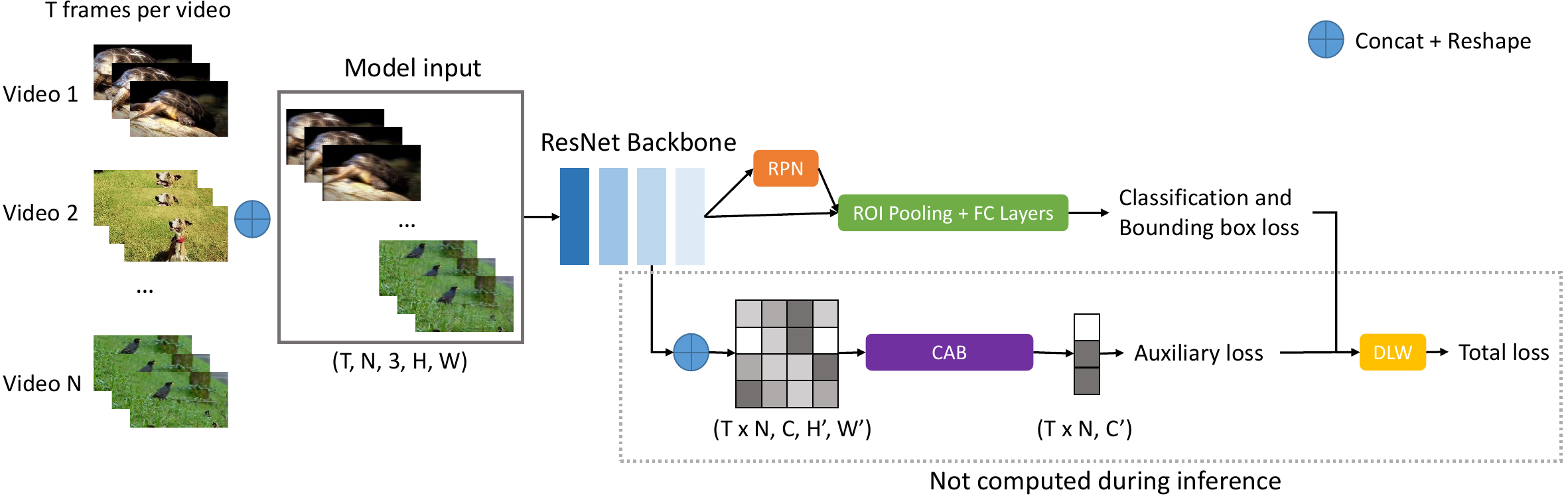}
\caption{\textbf{Overview of the Contrastive Learning through Auxiliary Branch (CLAB) framework.} We select $N$ videos, each containing $T$ extracted frames. During training, CAB takes as input an intermediate feature map generated by the backbone. Then, CAB projects this feature map into a 128-dimensional encoded feature vector, which is then used to compute the InfoNCE loss. The auxiliary loss is combined with the detection loss through the DLW module to compute the total loss. During testing, CAB and DLW are removed to improve detection performance without increasing computational cost.}
\label{overview}
\end{figure*}

Additionally, to efficiently integrate auxiliary knowledge into the model, we propose a dynamic loss weight strategy. This approach reduces the auxiliary loss weight throughout training, promoting the learning of general features in the early stages while ensuring the model focuses on the primary task as training progresses. An overview of our proposed method is shown in Figure \ref{overview}.

\subsection{Auxiliary Branch}

\begin{figure}[t]
\centering
\includegraphics[width=0.85\columnwidth]{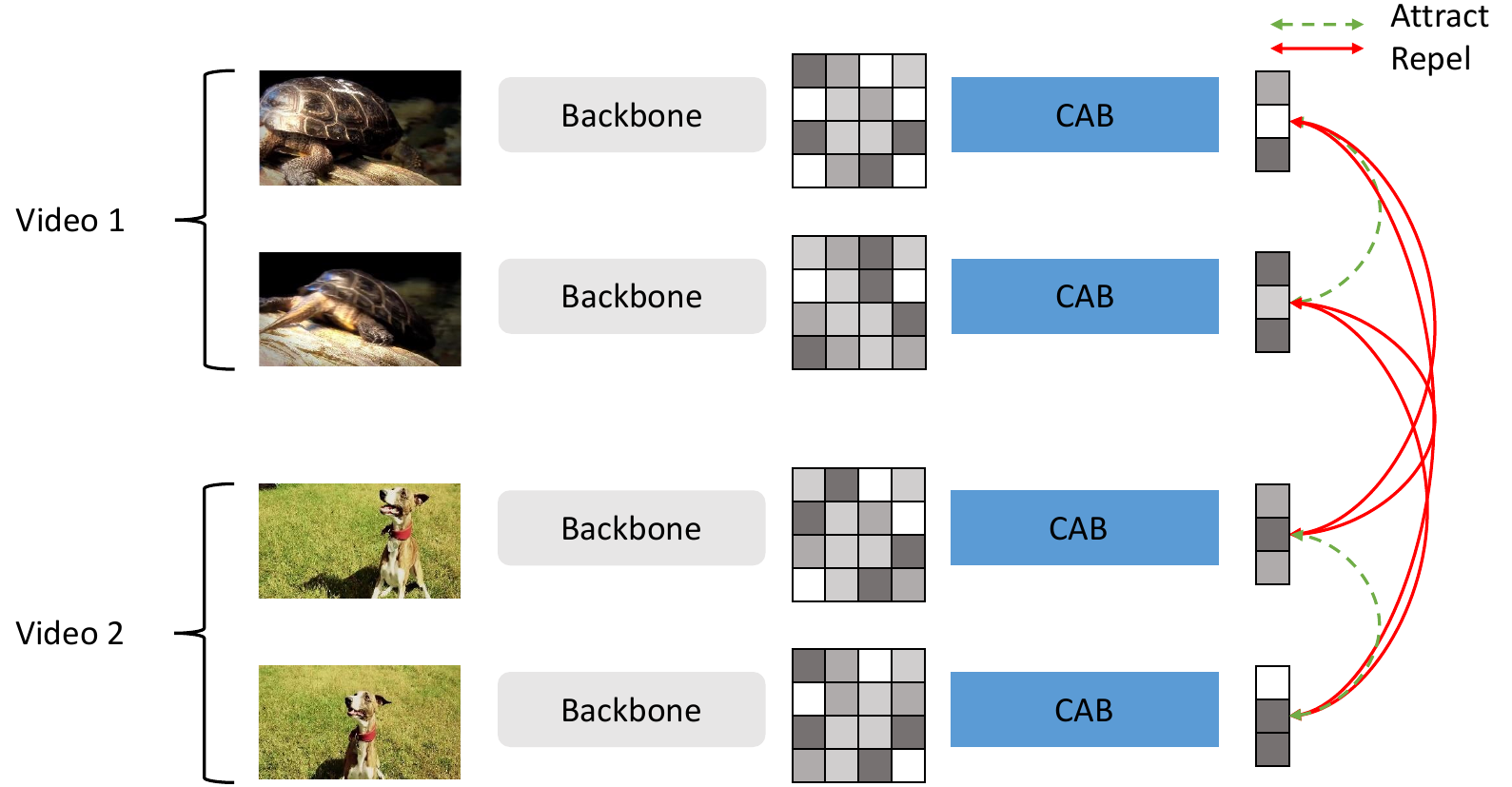}
\caption{Illustration of the InfoNCE loss in CLAB framework. The backbone and CAB are trained to pull together images from the same video and to push apart images from different videos, using the encoded feature vector and the contrastive auxiliary loss.}
\label{contrastive}
\end{figure}

Inspired by previous work in video representation learning \cite{qian2021spatiotemporal,dave2022tclr}, we introduce CAB, an auxiliary branch that applies the InfoNCE \cite{oord2018representation,he2020momentum,chen2020simple} contrastive loss on features extracted from video sequences. InfoNCE is considered a contrastive loss because it explicitly distinguishes between positive pairs, which are frames from the same video, and negative pairs, which are frames from different videos. By optimizing the loss, it encourages the model to bring features of frames from the same video sequence closer together in the feature space while pushing apart features from different videos. This process enhances robustness to frame degradation by improving the model's ability to distinguish relevant temporal patterns within video sequences. An illustration of the contrastive loss strategy is provided in Figure \ref{contrastive}.

As proposed by \cite{chen2017rethinking}, CAB takes an intermediate feature from the backbone to enhance robustness and mitigate vanishing gradients. The CAB architecture includes a convolutional layer with a 3 $\times$ 3 kernel, followed by a ReLU activation. The output is then processed by an adaptive average pooling layer, which reduces the spatial dimensions of the feature map to 1 $\times$ 1, meaning each channel is collapsed to a single average value, resulting in a compact, global representation of the input feature. The output is then flattened into a vector. Finally, as proposed by \cite{chen2020simple}, a projection module with two fully connected layers reduces the embedding dimensionality, with each layer followed by a ReLU activation, producing a 128-dimensional encoded feature vector used to compute the loss.

Suppose we have a batch of \( N \) videos, each containing \( T \) sampled frames. Let \( \mathbf{z}_{i} \) and \( \mathbf{z}_{i'} \) denote the encoded features of frames \( i \) and \( i' \) from the \( i \)-th video, respectively. The cosine similarity between two feature vectors \( u \) and \( v \) is defined by $\text{sim}(u, v) = \frac{u \cdot v}{\|u\| \|v\|}$.

The auxiliary InfoNCE contrastive loss, \( \mathcal{L}_{\text{aux}} \), is defined in Equation \ref{eq:infonce} as follows:

\begin{equation}
    \mathcal{L}_{\text{aux}}=-\frac{1}{N}\sum_{i=1}^{N}\log\frac{\exp(\text{sim}(\mathbf{z}_{i},\mathbf{z}_{i'})/\tau)}{\sum_{k=1}^{N \times T}\mathbf{1}_{[k\neq i]}\exp(\text{sim}(\mathbf{z}_{i},\mathbf{z}_{k})/\tau)}, \label{eq:infonce}
\end{equation}

Here, \( \tau \) is a temperature parameter, and \( \mathbf{1}_{[k \neq i]} \) is an indicator function that equals 1 if \( k \neq i \), and 0 otherwise. 

CAB is discarded during inference because it is an auxiliary branch that does not contribute to the final object detection. As it does not assist in the primary task, removing it leads to a more performant backbone without adding any extra computational overhead during evaluation.

\subsection{Dynamic Loss Weighting}

To enhance training stability using our auxiliary branch while maintaining focus on the primary task, we introduce a dynamic weighting strategy that integrates the auxiliary loss into the overall training objective. Building on insights from previous works on auxiliary tasks \cite{lin2019adaptive,du2018adapting}, we propose a straightforward yet effective approach called dynamic loss weighting.

As illustrated in Equation \ref{eq:dlw}, the auxiliary loss is initially assigned a weight \(w\), which decreases linearly over time, reaching zero at step \(k\).

\begin{equation}
w(t) = w \cdot \max\left(0, 1 - t / k \right), \label{eq:dlw}
\end{equation}

Here, \(t\) represents the current training step.

This schedule enables the auxiliary branch to contribute significantly to feature learning during the early stages of training. Over time, its influence diminishes, ensuring that the model prioritizes the primary task as training converges.

The auxiliary loss is then incorporated into the primary loss, as shown in Equation \ref{eq:loss}, where \(\mathcal{L}_{\text{regr}}\) and \(\mathcal{L}_{\text{classif}}\) represent the traditional detection losses (i.e., bounding box regression and object classification).

\begin{equation}
\mathcal{L}(t) = \mathcal{L}_{\text{regr}} + \mathcal{L}_{\text{classif}} + w(t) \mathcal{L}_{\text{aux}} \label{eq:loss}
\end{equation}

By leveraging this auxiliary loss, the model learns high-quality video representations, aligning similar images and improving robustness against frame degradation. Thanks to DLW strategy, the model benefits from these high-quality features during the initial stages of training while concentrating fully on video object detection as training concludes.

\section{Experiments}

Experiments use the ImageNet VID dataset \cite{deng2009imagenet} with 30 object categories for video object detection. The training set has 3,862 videos, and the validation set has 555. Following previous work, the models are trained using a combination of the ImageNet VID and ImageNet DET datasets, incorporating 30 classes from VID and the corresponding 30 overlapping classes from the 200 DET categories, and we use the split proposed by FGFA \cite{zhu2017flow}. Evaluation is done on the ImageNet VID validation set using mean average precision (mAP).

\subsection{Implementation Details}

The ResNet-101 \cite{he2016deep} (R101) is used as the backbone network for ablation studies. While we also use ResNeXt-101 \cite{xie2017aggregated} (X101) for the final results. 

Except for our CAB module, the model's architecture and hyperparameters follow those proposed by TROI \cite{gong2021temporal}. For a more detailed description of the architecture, we refer the reader to \cite{gong2021temporal}. CAB is composed by a convolutional layer with a 3 $\times$ 3 kernel, followed by a ReLU activation. The feature map is then processed by adaptive average pooling, which produces a 1 $\times$ 1 spatial map. This output is flattened and passed through a fully connected projection module consisting of two linear layers: the first layer maps from 512 to 256 dimensions, and the second maps from 256 to 128 dimensions, both followed by ReLU activations. The InfoNCE \cite{oord2018representation,he2020momentum,chen2020simple} loss is then computed on this sequence of vector with a temperature of $\tau = 0.1$. CAB is discarded during inference. In DLW, the auxiliary loss weight $w$ is initialized at 0.005 and gradually decreases to zero at $k=25,000$, corresponding to the midpoint of the training process.

The backbone networks are initialized with ImageNet pre-trained weights. Training is performed using SGD for 7 epochs with a total batch size of 16 across two A100 GPUs. The initial learning rate is 0.01, reduced by a factor of 10 at the 4th and 6th epochs. During training, one frame is sampled from the video along with 2 random support frames. Therefore, each image contains 2 positive pairs and 48 negative pairs. For inference, 30 support frames are sampled along with the target frame. Following STSN \cite{bertasius2018object}, if the support frames exceed the video boundaries, the first or last frame is repeated. Non-maximum suppression (NMS) with a 0.5 threshold is applied to remove duplicate detections. Images are resized to a shorter side of 600 pixels in both training and inference.

\subsection{Comparison With State-of-the-art Methods}

\begin{table}[t]
\caption{Comparisons with state-of-the-art methods on ImageNet VID validation set with various backbones (R101 / X101).}
\centering
\begin{tabular}{@{}c|cccccccc|c@{}}
    \toprule
    Backbone & FGFA  &  MANet & STSN  & SELSA  & TROI  & MEGA  & HVRNet  & BoxMask  & CLAB \\ \midrule \midrule
    R101 & 76.3 & 78.1 & 78.9 & 80.3 & 82.0 & 82.9 & 83.2 & 83.2 & \textbf{84.0} \\ 
    X101 & - & - & - & 83.1 & 84.3 & 84.1 & 84.8 & 84.8 & \textbf{85.2} \\
    \bottomrule[0.5pt]
\end{tabular}
\label{comparo}
\end{table}

While CLAB is based on TROI \cite{gong2021temporal}, it enables a direct performance comparison between CLAB and TROI. As shown in Table \ref{comparo}, CLAB outperforms TROI by 2.0 points and 0.9 points for the ResNet-101 \cite{he2016deep} and ResNeXt-101 \cite{xie2017aggregated} backbones, respectively, without adding any computational cost during inference.

Table \ref{comparo} further highlights the performance of the proposed method alongside other state-of-the-art models on the ImageNet VID validation set. The compared methods span various categories: motion-based methods like FGFA \cite{zhu2017flow}, MANet \cite{wang2018fully} and STSN \cite{bertasius2018object}; attention-based methods such as SELSA \cite{wu2019sequence}, TROI \cite{gong2021temporal}, and BoxMask \cite{hashmi2023boxmask}; and relation-based methods including MEGA \cite{chen2020memory} and HVRNet \cite{han2020mining}. For fairness, all methods were evaluated without using postprocessing-based scoring techniques such as Seq-NMS \cite{han2016seq} or BLR \cite{deng2019relation}. Additionally, only CNN-based methods were included, so results from \cite{wang2022ptseformer,roh2023diffusionvid} are not presented.

Without bells and whistles, CLAB achieves 84.0\% mAP with ResNet-101, surpassing well-known motion-based methods like STSN, as well as optical flow-based approaches such as FGFA and MANet. SELSA and TROI are simplified versions of our approach and therefore naturally achieve lower performance than CLAB. Notably, we successfully outperform state-of-the-art methods such as MEGA, HVRNet, and BoxMask while maintaining a significantly simpler design. Using a stronger ResNeXt-101 backbone, CLAB achieved an 85.2\% mAP, marking the highest accuracy among competitors and demonstrating the effectiveness of CLAB method in enhancing robustness and overall performance.

\subsection{Ablation Study}

\textbf{CAB and DLW} We evaluate the effectiveness of our proposed modules, CAB and DLW, as summarized in Table \ref{Ablation-module}. Starting with the TROI \cite{gong2021temporal} baseline, we incrementally incorporate CAB and DLW into CLAB method to validate the contributions of each module. First, integrating CAB alone results in a 1.4\% improvement in mAP, highlighting the benefits of contrastive learning for producing high-quality video representations. Next, adding DLW further enhances the performance, yielding an additional 0.6\% gain, thanks to the effective integration of contrastive knowledge without disrupting the primary task. Overall, the combination of CAB and DLW achieves a total performance boost of 2\% compared to the baseline, all without introducing any additional computational overhead during inference.

\begin{table}[t]
\caption{Ablation studies of CAB and DLW.}
	\centering
	\begin{tabular}{@{}cccc@{}}
		\toprule
		Method           & CAB          & DLW & mAP\! (\%)  \\ \midrule
		Temporal ROI Baseline \cite{gong2021temporal} &                  &               & 82.0   \\
		CLAB & \checkmark              &               & 83.4   \\
		CLAB & \checkmark                  & \checkmark              & \textbf{84.0}   \\
		 \bottomrule[0.5pt]
	\end{tabular}
	\label{Ablation-module}
\end{table}

\begin{table}[t]
    \centering
    \begin{minipage}{0.45\textwidth}
        \centering
\caption{Ablation on temperature.}
	\centering
	\begin{tabular}{@{}ccccc@{}}
		\toprule
		$\tau$           & 0.05          & 0.1 & 0.5 & 1  \\ \midrule
		mAP\! (\%) &  83.1                & \textbf{84.0}       & 83.5       & 83.6  \\
		 \bottomrule[0.5pt]
	\end{tabular}
	\label{Ablation-temperature}
    \end{minipage}
    \hfill
    \begin{minipage}{0.45\textwidth}
\caption{Ablation on loss weight.}
	\centering
	\begin{tabular}{@{}ccccc@{}}
		\toprule
		$w$           & 0.001          & 0.005 & 0.01 & 0.05  \\ \midrule
		mAP\! (\%) &  83.1                & \textbf{84.0}       & 83.1       & 82.7  \\
		 \bottomrule[0.5pt]
	\end{tabular}
	\label{Ablation-weight}
    \end{minipage}
\end{table}

\begin{figure*}[t]
\centering
\includegraphics[width=1.0\textwidth]{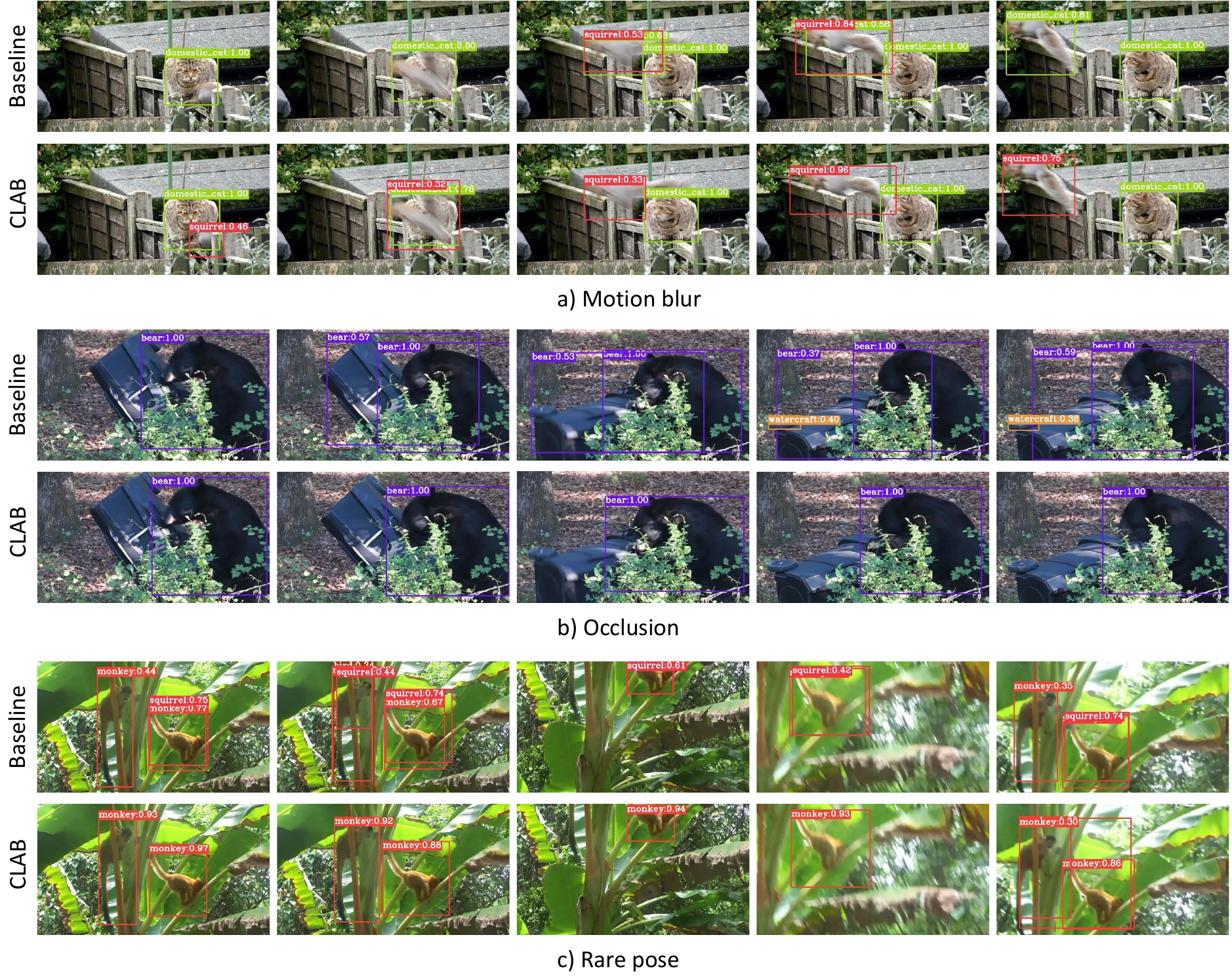}
\caption{Visualization of detection results for various types of typical image degradation due to video. For each video, the first row shows the TROI \cite{gong2021temporal} baseline, while the second row displays the results of CLAB method.}
\label{visualisation}
\end{figure*}

\textbf{Temperature} As proposed by \cite{chen2020simple,he2020momentum}, the temperature hyperparameter $\tau$ of the InfoNCE loss plays a critical role in model performance. It controls the sharpness of the softmax distribution, balancing the weighting of positive and negative pairs. A small temperature ($\tau < 0.1$) emphasizes hard negatives and sharpens decision boundaries but can lead to an instable training, while a large temperature ($\tau > 1$) reduces contrast, treating positives and negatives too similarly and lowering discriminative power. As shown in Table \ref{Ablation-temperature}, a temperature of 0.1 achieves the best performance with 84\% mAP and is chosen as the default. Increasing $\tau$ to 0.5 or 1 slightly reduces performance by 0.5\% and 0.4\% mAP, respectively, whereas lowering $\tau$ to 0.05 significantly drops performance by nearly 1\%. This indicates that for auxiliary tasks, higher temperatures are preferable to stabilize training.

\textbf{Auxiliary loss weight} Table \ref{Ablation-weight} shows the impact of the auxiliary loss weight on performance. We found that a weight $w$ of 0.005 yields the best performance and is set as the default. Increasing or decreasing this weight leads to performance degradation. Nevertheless, CLAB consistently outperforms the baseline TROI, achieving 82.7\% mAP even with a loss weight of 0.05 compared to 82.0\% mAP for TROI, demonstrating the effectiveness of our approach.

\textbf{Visualisation results} Figure \ref{visualisation} presents the detection results for several video clips, arranged in chronological order. We compare our proposed CLAB method (second row) with TROI (first row), as CLAB is built upon TROI, a previous state-of-the-art approach. In snippet (a), the squirrel initially appears in an unusual pose, making it challenging to recognize. Subsequently, its rapid movements result in motion blur. In snippet (b), the bear is partially obscured by a bush, while the trash can overlaps with the bear. In snippet (c), the monkeys start in rare poses, and the camera’s rapid movement introduces motion blur. 

Comparing the detection results of TROI and CLAB clearly shows that CLAB significantly improves detection quality. In snippet (a), CLAB consistently detects the squirrel throughout the sequence, whereas TROI only identifies it when there is no overlap with the cat and assigns a low confidence score when it does. In snippet (b), while both methods confidently detect the bear despite obstruction, TROI misclassifies a trash can as another bear or even a watercraft. In snippet (c), TROI struggles with a monkey in a rare pose, often misidentifying it as a squirrel and assigning a low confidence score to another monkey. In contrast, CLAB reliably detects both monkeys, regardless of pose or camera movement.

\section{Conclusion}

In this paper, we propose CLAB, a simple yet effective approach to improve video object detection performance without adding any computational cost at inference. Our method introduces an auxiliary branch with a contrastive loss to bring together features from the same video in the feature space. We combine this auxiliary branch with a dynamic loss weighting strategy to effectively integrate the auxiliary knowledge into the model. Extensive experiments on ImageNet VID demonstrate that CLAB can effectively boost performance and surpass state-of-the-art CNN-based methods. 
We hope this work contributes to advancing video object detection while maintaining reduced computational requirements.

\bibliographystyle{splncs04}
\bibliography{clab}

\begin{thebibliography}{10}
\providecommand{\url}[1]{\texttt{#1}}
\providecommand{\urlprefix}{URL }
\providecommand{\doi}[1]{https://doi.org/#1}

\bibitem{araslanov2020single}
Araslanov, N., Roth, S.: Single-stage semantic segmentation from image labels. In: Proceedings of the IEEE/CVF conference on computer vision and pattern recognition. pp. 4253--4262 (2020)

\bibitem{bertasius2018object}
Bertasius, G., Torresani, L., Shi, J.: Object detection in video with spatiotemporal sampling networks. In: Proceedings of the European Conference on Computer Vision (ECCV). pp. 331--346 (2018)

\bibitem{chen2017rethinking}
Chen, L.C.: Rethinking atrous convolution for semantic image segmentation. arXiv preprint arXiv:1706.05587  (2017)

\bibitem{chen2020simple}
Chen, T., Kornblith, S., Norouzi, M., Hinton, G.: A simple framework for contrastive learning of visual representations. In: International conference on machine learning. pp. 1597--1607. PMLR (2020)

\bibitem{chen2020memory}
Chen, Y., Cao, Y., Hu, H., Wang, L.: Memory enhanced global-local aggregation for video object detection. In: Proceedings of the IEEE/CVF conference on computer vision and pattern recognition. pp. 10337--10346 (2020)

\bibitem{cui2022mixformer}
Cui, Y., Jiang, C., Wang, L., Wu, G.: Mixformer: End-to-end tracking with iterative mixed attention. In: Proceedings of the IEEE/CVF Conference on Computer Vision and Pattern Recognition. pp. 13608--13618 (2022)

\bibitem{dave2022tclr}
Dave, I., Gupta, R., Rizve, M.N., Shah, M.: Tclr: Temporal contrastive learning for video representation. Computer Vision and Image Understanding  \textbf{219},  103406 (2022)

\bibitem{deng2009imagenet}
Deng, J., Dong, W., Socher, R., Li, L.J., Li, K., Fei-Fei, L.: Imagenet: A large-scale hierarchical image database. In: 2009 IEEE conference on computer vision and pattern recognition. pp. 248--255. Ieee (2009)

\bibitem{deng2019relation}
Deng, J., Pan, Y., Yao, T., Zhou, W., Li, H., Mei, T.: Relation distillation networks for video object detection. In: Proceedings of the IEEE/CVF international conference on computer vision. pp. 7023--7032 (2019)

\bibitem{du2018adapting}
Du, Y., Czarnecki, W.M., Jayakumar, S.M., Farajtabar, M., Pascanu, R., Lakshminarayanan, B.: Adapting auxiliary losses using gradient similarity. arXiv preprint arXiv:1812.02224  (2018)

\bibitem{gong2021temporal}
Gong, T., Chen, K., Wang, X., Chu, Q., Zhu, F., Lin, D., Yu, N., Feng, H.: Temporal roi align for video object recognition. In: Proceedings of the AAAI Conference on Artificial Intelligence. vol.~35, pp. 1442--1450 (2021)

\bibitem{han2020mining}
Han, M., Wang, Y., Chang, X., Qiao, Y.: Mining inter-video proposal relations for video object detection. In: Computer Vision--ECCV 2020: 16th European Conference, Glasgow, UK, August 23--28, 2020, Proceedings, Part XXI 16. pp. 431--446. Springer (2020)

\bibitem{han2016seq}
Han, W., Khorrami, P., Paine, T.L., Ramachandran, P., Babaeizadeh, M., Shi, H., Li, J., Yan, S., Huang, T.S.: Seq-nms for video object detection. arXiv preprint arXiv:1602.08465  (2016)

\bibitem{hashmi2023boxmask}
Hashmi, K.A., Pagani, A., Stricker, D., Afzal, M.Z.: Boxmask: Revisiting bounding box supervision for video object detection. In: Proceedings of the IEEE/CVF Winter Conference on Applications of Computer Vision. pp. 2030--2040 (2023)

\bibitem{he2020momentum}
He, K., Fan, H., Wu, Y., Xie, S., Girshick, R.: Momentum contrast for unsupervised visual representation learning. In: Proceedings of the IEEE/CVF conference on computer vision and pattern recognition. pp. 9729--9738 (2020)

\bibitem{he2016deep}
He, K., Zhang, X., Ren, S., Sun, J.: Deep residual learning for image recognition. In: Proceedings of the IEEE conference on computer vision and pattern recognition. pp. 770--778 (2016)

\bibitem{lin2019adaptive}
Lin, X., Baweja, H., Kantor, G., Held, D.: Adaptive auxiliary task weighting for reinforcement learning. Advances in neural information processing systems  \textbf{32} (2019)

\bibitem{Liu_2023_ICCV}
Liu, X., Nejadasl, F.K., van Gemert, J.C., Booij, O., Pintea, S.L.: Objects do not disappear: Video object detection by single-frame object location anticipation. In: Proceedings of the IEEE/CVF International Conference on Computer Vision (ICCV). pp. 6950--6961 (October 2023)

\bibitem{oord2018representation}
Oord, A.v.d., Li, Y., Vinyals, O.: Representation learning with contrastive predictive coding. arXiv preprint arXiv:1807.03748  (2018)

\bibitem{qian2021spatiotemporal}
Qian, R., Meng, T., Gong, B., Yang, M.H., Wang, H., Belongie, S., Cui, Y.: Spatiotemporal contrastive video representation learning. In: Proceedings of the IEEE/CVF conference on computer vision and pattern recognition. pp. 6964--6974 (2021)

\bibitem{ren2015faster}
Ren, S.: Faster r-cnn: Towards real-time object detection with region proposal networks. arXiv preprint arXiv:1506.01497  (2015)

\bibitem{roh2023diffusionvid}
Roh, S.D., Chung, K.S.: Diffusionvid: Denoising object boxes with spatio-temporal conditioning for video object detection. IEEE Access  (2023)

\bibitem{russakovsky2015imagenet}
Russakovsky, O., Deng, J., Su, H., Krause, J., Satheesh, S., Ma, S., Huang, Z., Karpathy, A., Khosla, A., Bernstein, M., et~al.: Imagenet large scale visual recognition challenge. International journal of computer vision  \textbf{115},  211--252 (2015)

\bibitem{tzeng2015simultaneous}
Tzeng, E., Hoffman, J., Darrell, T., Saenko, K.: Simultaneous deep transfer across domains and tasks. In: Proceedings of the IEEE international conference on computer vision. pp. 4068--4076 (2015)

\bibitem{wang2022ptseformer}
Wang, H., Tang, J., Liu, X., Guan, S., Xie, R., Song, L.: Ptseformer: Progressive temporal-spatial enhanced transformer towards video object detection. In: European Conference on Computer Vision. pp. 732--747. Springer (2022)

\bibitem{wang2018fully}
Wang, S., Zhou, Y., Yan, J., Deng, Z.: Fully motion-aware network for video object detection. In: Proceedings of the European conference on computer vision (ECCV). pp. 542--557 (2018)

\bibitem{wu2019sequence}
Wu, H., Chen, Y., Wang, N., Zhang, Z.: Sequence level semantics aggregation for video object detection. In: Proceedings of the IEEE/CVF international conference on computer vision. pp. 9217--9225 (2019)

\bibitem{xie2017aggregated}
Xie, S., Girshick, R., Doll{\'a}r, P., Tu, Z., He, K.: Aggregated residual transformations for deep neural networks. In: Proceedings of the IEEE conference on computer vision and pattern recognition. pp. 1492--1500 (2017)

\bibitem{xu2020knowledge}
Xu, G., Liu, Z., Li, X., Loy, C.C.: Knowledge distillation meets self-supervision. In: European conference on computer vision. pp. 588--604. Springer (2020)

\bibitem{zhai2019s4l}
Zhai, X., Oliver, A., Kolesnikov, A., Beyer, L.: S4l: Self-supervised semi-supervised learning. In: Proceedings of the IEEE/CVF international conference on computer vision. pp. 1476--1485 (2019)

\bibitem{zhang2021bytetrack}
Zhang, Y., Sun, P., Jiang, Y., Yu, D., Yuan, Z., Luo, P., Liu, W., Wang, X.: Bytetrack: Multi-object tracking by associating every detection box (2021)

\bibitem{zhao2017pyramid}
Zhao, H., Shi, J., Qi, X., Wang, X., Jia, J.: Pyramid scene parsing network. In: Proceedings of the IEEE conference on computer vision and pattern recognition. pp. 2881--2890 (2017)

\bibitem{zhu2017flow}
Zhu, X., Wang, Y., Dai, J., Yuan, L., Wei, Y.: Flow-guided feature aggregation for video object detection. In: Proceedings of the IEEE international conference on computer vision. pp. 408--417 (2017)

\bibitem{zhuang2020unsupervised}
Zhuang, C., She, T., Andonian, A., Mark, M.S., Yamins, D.: Unsupervised learning from video with deep neural embeddings. In: Proceedings of the ieee/cvf conference on computer vision and pattern recognition. pp. 9563--9572 (2020)

\end{thebibliography}

\end{document}